\documentclass[10pt,twocolumn,letterpaper]{article}

\usepackage{cvpr}
\usepackage{times}
\usepackage{epsfig}
\usepackage{graphicx}
\usepackage{amsmath}
\usepackage{amssymb}
\usepackage{enumitem}

\usepackage{multirow}
\usepackage{multicol}

\usepackage[pagebackref=true,breaklinks=true,letterpaper=true,colorlinks,bookmarks=false]{hyperref}

\newcommand{\obj}{\mathrm{Obj}}
\newcommand{\conf}{\mathit{conf}}
\newcommand{\Conf}{\mathrm{Conf}}
\newcommand{\Attr}{\mathrm{Attr}}
\newcommand{\Clf}{\mathrm{Clf}}
\newcommand{\attr}{\mathrm{attr}}
\newcommand{\clf}{\mathrm{clf}}
\newcommand{\argmin}{\arg\min}

\cvprfinalcopy 

\def\cvprPaperID{5257} 

\ifcvprfinal\fi
\begin{document}
\setlength{\abovedisplayskip}{3pt}
\setlength{\belowdisplayskip}{3pt}
\title{Don't Even Look Once: Synthesizing Features for Zero-Shot Detection}

\author{Pengkai~Zhu$^\dagger$ \hspace{4em} Hanxiao~Wang$^{\dagger,\ddagger}$ \hspace{2em} Venkatesh Saligrama$^\dagger$\\
$^\dagger$ Electrical and Computer Engineering Department, Boston University\\
$^\ddagger$ Onfido ltd, London, UK\\
{\tt\small $^\dagger$\{zpk, hxw, srv\}@bu.edu, $^\ddagger$hanxiao.wang@onfido.com}
}

\maketitle

\begin{abstract}
\vspace{-1mm}
Zero-shot detection, namely, localizing both seen and unseen objects, increasingly gains importance for large-scale applications, with large number of object classes, since, collecting sufficient annotated data with ground truth bounding boxes is simply not scalable. While vanilla deep neural networks deliver high performance for objects available during training, unseen object detection degrades significantly. At a fundamental level, while vanilla detectors are capable of proposing bounding boxes, which include unseen objects, they are often incapable of assigning high-confidence to unseen objects, due to the inherent precision/recall tradeoffs that requires rejecting background objects. We propose a novel detection algorithm ``Don't Even Look Once (DELO),'' that synthesizes visual features for unseen objects and augments existing training algorithms to incorporate unseen object detection. Our proposed scheme is evaluated on PascalVOC and MSCOCO, and we demonstrate significant improvements in test accuracy over vanilla and other state-of-art zero-shot detectors.
\end{abstract}





\vspace{-5.5mm}
\section{Introduction\label{sec.intro}}
\vspace{-1mm}
While deep learning based object detection methods have achieved impressive average precision over the last five years~\cite{girshick2015fast,ren2015faster,redmon2016you,liu2016ssd,he2017mask,redmon2016yolo9000}, these gains can be attributed to the availability of training data in the form of fully annotated ground-truth object bounding boxes. 

\noindent
\textbf{Zero-Shot Detection (ZSD): The Need.} As we scale up detection to large-scale applications and ``in the wild'' scenarios, the demand for bounding-box level annotations across a large number of object classes is simply not scalable. 
Consequently, as object detection moves towards large-scale
\footnote{ Although annotations have increased in common detection datasets (e.g. 20 classes provided by {\it PASCAL VOC}~\cite{everingham2010pascal}; 80 in {\it MSCOCO}~\cite{lin2014microsoft}), the size is substantially smaller relative to image classification~\cite{deng2009imagenet}.}
, it is imperative that we move towards a framework that serves the dual role of detecting objects seen during training as well as detecting unseen classes as and when they appear at test-time.

\noindent \textbf{Reusing Existing Detectors.} Vanilla DNN detectors relegate {\it unseen} objects into the background leading to missed detection of unseen objects. To understand the root of this issue, we note that most detectors, base their detection, on three components, (a) proposing object bounding boxes; (b) outputting objectness score to provide confidence for a candidate bounding box, and to filter out bounding boxes with low confidence; (c) a classification score for recognizing the object in a high-confidence bounding box. 

\noindent \textbf{Objectness Scores.} Evidently, our empirical results suggest that, of the three different components, the high miss-detection rate of vanilla DNN detectors for unseen objects can be attributed to (b). Indeed, (a) is less of an issue, since existing detectors typically propose hundreds of bounding boxes per image, which also include unseen objects, but are later filtered out because of poor objectness scores. 
Finally, (c) is also not a significant issue, since, conditioned on having a good bounding box, the classification component performs sufficiently well even for unseen objects with rates approaching zero-shot recognition accuracy (i.e., classification with ground-truth bounding boxes). Consequently, the performance loss primarily stems from assigning poor confidence to bounding boxes that do not contain seen objects. On the other hand, naively modifying confidence penalty, while improving recall, leads to poor precision, as the system tends to assign higher confidence to bounding boxes that are primarily part of the background as well. 

\noindent \textbf{Novelty.} We seek to improve confidence predictions on bounding boxes with sufficient overlap with seen and unseen objects, while still ensuring low confidence on bounding boxes that primarily contain background. Our dilemma is that we do not observe unseen objects during training, even possibly as unannotated images. With this in mind, we propose to leverage semantic vectors of unseen objects, and construct synthetic unseen features based on a conditional variational auto-encoder (CVAE). To train a confidence predictor, we then propose to augment the current training pipeline, composed of the three components outlined above, along with the unseen synthetic visual features. This leads to a modified empirical objective for confidence prediction that seeks to assign higher confidence to bounding boxes that bear similarity to the synthesized unseen features as well as real seen features, while ensuring low confidence on bounding boxes that primarily contain the background. In addition, we propose a sampling scheme, whereby during training, the proposed bounding boxes are re-sampled so as to maintain a balance between background and foreground objects. Our scheme is inspired by focal loss~\cite{lin2017focal}, and seeks to overcome the significant foreground-background imbalance, which tends to reduce recall, and in particular adversely impacts unseen classification.  

\noindent
\textbf{Evaluation.} ZSD algorithms must be evaluated carefully to properly attribute gains to the different system components. For this reason we list four principal attributes that are essential for validating performance in this context:\\
\textbf{(a)} \underline{Dataset Complexity.} Datasets such as ImageNet~\cite{deng2009imagenet} typically contain one object/image; and F-MNIST~\cite{xiao2017fashion} in addition has a dark background. As such, detection is somewhat straightforward obviating the need to employ DNNs. For this reason, we consider only those datasets containing multiple objects per image such as MSCOCO~\cite{lin2014microsoft} and PascalVOC~\cite{everingham2010pascal}, where DNN detectors are required to realize high precision.\\
\textbf{(b)} \underline{Protocol.} During training we admit images that contain only seen class objects, and filter out any image containing unseen objects (so transductive methods are omitted in our comparison). We follow \cite{zhu2019zero} and consider three sets of evaluations: Test-seen (TS), Test-Unseen (TU) and Test-Mix (TM). The goal of test-seen is to benchmark performance of proposed method against vanilla detectors, which are optimal for this task. The goal of test-unseen evaluation is to evaluate performance when only unseen objects are present, which is analogous to the purely zero-shot evaluation in the recognition context~\cite{xian2018zero}. Test-mix containing a mixture of both seen and unseen objects typically within the same image is the most challenging, and can be viewed analogous to generalized zero-shot setting. \\
\textbf{(c)} \underline{High vs. Low Seen-to-Unseen Splits.} The number of objects seen during training vs. test-time determines the efficacy of the detection algorithm. At high seen/unseen object class ratios, evidently, gains are predominantly a function of recognition algorithm, necessitating no improvement object bounding boxes on unseen objects. For this reason, we experiment with a number of different splits. \\
\textbf{(d)} \underline{AP and mAP.} Once a bounding box is placed around a valid object, the task of 
recognition can usually be performed by passing the bounding box through any ZSR algorithm. As such mAP performance gain could be credited to improvements in placing high-confidence bounding boxes (as reflected by AP) in the right places as well as improvements in ZSR algorithm. For instance, as we noted above, high seen/unseen ratios can be attributed to improved recognition. For this reason we tabulate APs for different splits.

\begin{figure*}[t]
    \centering
    \includegraphics[width=\textwidth]{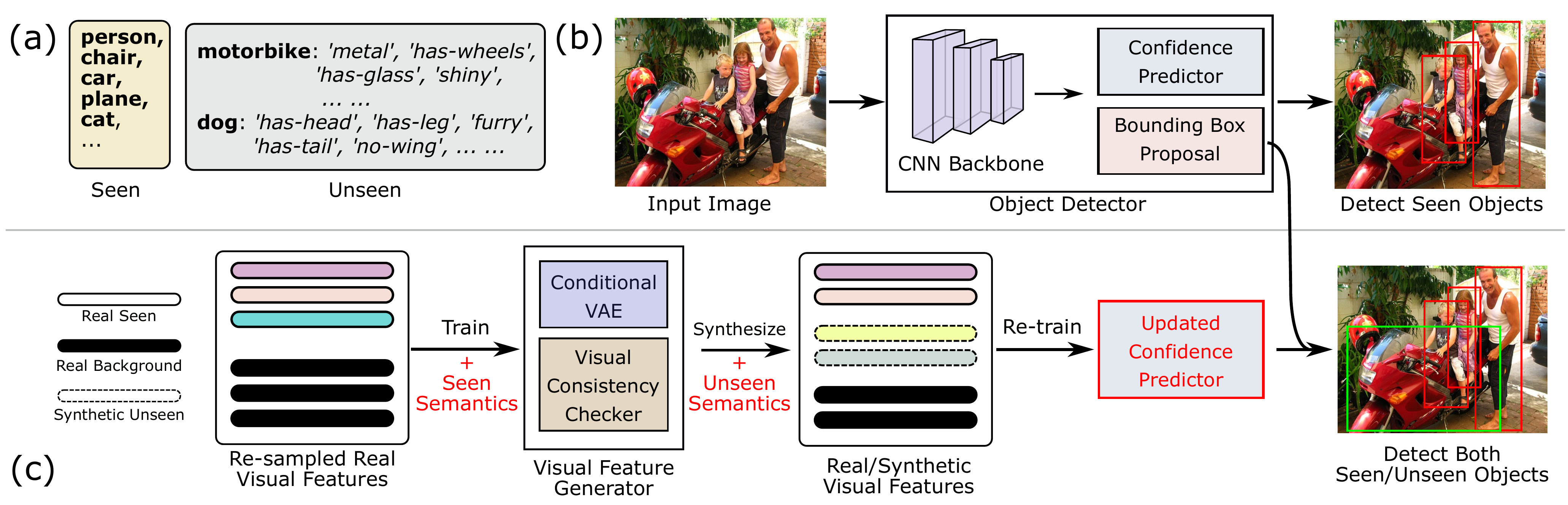}
    \caption{
    (a) An illustration of seen/unseen classes and the semantic description;
    (b) A vanilla detector trained using seen objects only tends 
    to relegate the confidence score of unseen objects;
    (c) The proposed approach. We first train a visual feature generator
    by taking a pool of visual features of foreground/background objects and their semantics with a balanced ratio. We then use it to synthesize visual features for unseen objects;
    Finally we add the synthesized visual features back to the pool
    and re-train the confidence predictor module of the vanilla detector.
    The re-trained confidence predictor can be plugged back into the detector and detect unseen objects.
    }
    \label{fig:overview}
    \vspace{-5mm}
\end{figure*}

\vspace{-1mm}
\section{Related Work\label{sec.related}}
\vspace{-1mm}

\noindent \textbf{Traditional vs. Generalized ZSL (GZSL)}. 
Zero Shot Learning (ZSL) seeks to recognize novel visual categories that are unannotated in training data \cite{lampert2014attribute,Xian_2016_CVPR,lei2015predicting,zhang2016zero}. As such, ZSL exhibits bias towards unseen classes, and GZSL evaluation attempts to rectify it by evaluating on both seen and unseen objects at test-time \cite{Chao2016,xian2017zero,fu2017recent,Jiang_2019_ICCV,Huang_2019_CVPR, Zhu_2019_CVPR,Schonfeld_2019_CVPR,Zhu_2019_ICML}. Our evaluation for ZSD follows GZSL focusing on both seen and unseen objects.  

\noindent \textbf{Generative ZSL methods}.
Semantic information is a key ingredient for transferring knowledge from seen to unseen classes. This can be in the form of attributes~\cite{farhadi2009describing,lampert2014attribute,mensink2012metric,parikh2011interactively, castanon2016retrieval, chen2018probabilistic}, word phrases~\cite{socher2013zero,frome2013devise}, etc. Such semantic data is often easier to collect and the premise of many ZSL methods is to substitute hard to collect visual samples for semantic data. Nevertheless, there is often a large visual-semantic gap, which results in significant performance degradation. Motivated by these concerns, recent works have proposed to synthesize unseen examples by means of generative models such as autoencoders~\cite{chen2018zero,kodirov2017semantic}, GANs and adversarial methods~\cite{zhu2018generative,Verma_2018_CVPR,Xian_2018_CVPR,huang2019generative,li2019leveraging,sariyildiz2019gradient} that take semantic vectors as input, and output images. Following their approach, we propose to similarly bridge visual-semantic gap in ZSD through synthesizing visual features for unseen objects (since visual images are somewhat noisy). 

\noindent \textbf{Zero-Shot Detection}.
Recently, a few papers have begun to focus attention on zero-shot detection~\cite{bansal2018zero, rahman2018zero,li2019zero,rahman2019transductive,zhu2019zero}. Unfortunately, methods, datasets, protocols and splits are all somewhat different, and the software code is not publicly available to comprehensively validate against all the methods. Nevertheless, we will highlight here some of the differences within the context of our evaluation metric (a-d). 

First, \cite{rahman2018zero,li2019zero} evaluate on Test-Unseen (TU) problem. Analogous to ZSL vs. GZSL, optimizing for TU can induce an unseen class bias resulting in poor performance on seen. Furthermore, \cite{bansal2018zero} tabulate GZSD performance (because purportedly mAP is low) only as a recall rate wrt top-100 bounding boxes ranked according to confidence scores. As such, there are fewer than 10 foreground objects per image, this metric is difficult to justify, since 100 bounding boxes typically includes all unseen objects. \cite{rahman2019transductive} proposes a transductive approach, which while evaluates seen and unseen objects, leverages appearance of unseen images during training. In contrast to these works, and like \cite{zhu2019zero} we evaluate our method in the full GZSD setting.

Second, methodologically these works ~\cite{bansal2018zero, rahman2018zero,li2019zero,rahman2019transductive} could be viewed as contributing to post-processing of detected bounding boxes, leveraging extensions to zero-shot recognition systems. In essence, these methods take outputs of existing vanilla detectors as given (region proposal network (RPN)~\cite{rahman2018zero,rahman2019transductive,li2019zero} or Edge-Box~\cite{bansal2018zero}), and take their bounding boxes along with confidence scores as inputs into their system. This means that their gains arise primarily from improved recognition rather than in placing bounding boxes of high-confidence. In contrast, and like \cite{zhu2019zero}, we attempt to improve localization performance by outputting high-confidence bounding boxes for unseen objects. Nevertheless, unlike \cite{zhu2019zero}, who primarily utilize semantic and visual vectors of seen classes to improve confidence of bounding boxes, we synthesize unseen visual features as well. As a result, we outperform \cite{zhu2019zero} in our various evaluations.

Third, there is an issue of complexity of datasets, and as to how they are evaluated. \cite{Demirel2018ZeroShotOD} tabulates F-MNIST with clear black backgrounds, and \cite{rahman2018zero} ImageNet with only single object/image, both of which are not informative from a detection perspective. Then, there is the issue of splits. A number of these methods exclusively consider high seen/unseen object class ratio splits ($16/4$ in \cite{Demirel2018ZeroShotOD}, $177/23$ in \cite{rahman2018zero,rahman2019transductive}, and $48/17$ and $478/130$ in \cite{bansal2018zero}). Such high split ratios could be uninformative, since we maybe in a situation where the visual features of the seen class could be quite similar to the unseen class, resulting in placing sufficiently large number of bounding boxes on unseen objects. This coupled with recall@100 metric or TU evaluations could exhibit unusually high gains. Finally, AP scores are seldom tabulated, which from our viewpoint would be informative about localization performance. In contrast, and following \cite{zhu2019zero} we consider several splits, different metrics (AP, mAP, recall@100) and tabulate performance on detection datasets such as MSCOCO and PascalVOC.

\vspace{-2mm}
\section{Methodology\label{sec.method}}
\noindent \textbf{Problem Definition.}
We formally define zero-shot detection (ZSD). A training dataset of $M$ images with corresponding objects labels $\mathcal{D}_{tr} = \{I_m, \{\obj_m^{(i)}\}_{i=1}^{N_m}\}_{m=1}^M$ is provided, where $N_m$ is the number of objects and $\{\obj_m^{(i)}\}_{i=1}^{N_m}$ is the collection of all objects labels in the image $I_m$. Every object is labeled as $\obj = \{{\bf B}, c\}$ where ${\bf B} = \{x,y,w,h\}$ is the location and the size of the bounding box and $c \in {\bf C}_{seen}$ is the class label (sup-/subscript are omitted when clear). For testing, images containing objects from both seen (${\bf C}_{seen}$) and unseen classes (${\bf C}_{unseen}$) are given with ${\bf C}_{seen} \cap {\bf C}_{unseen} = \varnothing$). The task is to predict the bounding box for every foreground objects. Additionally, for training the semantic representation $S_c$ of all classes ($c \in {\bf C}_{all} = {\bf C}_{seen} \cup {\bf C}_{unseen}$) are also provided. \\

\noindent \textbf{Backbone Architecture.}
We use YOLOv2\cite{redmon2017yolo9000} as a baseline. However, our proposed method can readily incorporate single stage detectors (SSD) or region-proposal-network (RPN). We briefly describe YOLOv2 below.


YOLOv2 is a fully convolutional network and consists of two modules: a feature extractor  $F$ and an object predictor. The feature extractor is implemented by Darknet-19~\cite{redmon2016you}, which takes input image size $416 \times 416$ and outputs the convolutional feature maps $F(I_m)$ with size $13 \times 13 \times 1024$. The object predictor is implemented by a $1 \times 1$ convolutional layer, which contains 5 bounding boxes predictors assigned with 5 anchor boxes with predefined aspect ratios for prediction diversity. Each bounding box predictor consists of an object locator, which outputs the bonding box location and size $\hat{{\bf B}}$, a confidence predictor $\Conf$, which outputs the objectness score $p_\conf$ of the bounding box, and a classifier. The objectness score is in $[0, 1]$ and denotes the confidence of whether the bounding box contains foreground object (1) or background (0). The bounding boxes predictors convolves on every cell of $F(I_m)$ and make detection predictions for the entire image.

\subsection{System Overview\label{subsec.procedure}}
\noindent The three objectives in our context are:
\textbf{(1)} Improve {\textit{bad precision-recall for unseen objects}} whose confidence scores are suppressed by detectors trained with seen classes;
\textbf{(2)} Deal with {\textit{background/foreground imbalance}} that hampers the precision; and
\textbf{(3)} Account for {\textit{generalized ZSD performance}} where both seen/unseen objects exist in the test set.

\noindent \textbf{Key Idea}.
All of these objectives can be realized by improving the confidence predictor component, whereby both seen and unseen object bounding boxes receive higher confidence while background objects are still suppressed. To do so we retrain confidence predictor by leveraging real visual features for seen and background objects, and synthetic features for unseen objects. We resample bounding boxes to correct the background/foreground imbalance. 
Fig.\ref{fig:overview} depicts the four stages of the proposed pipeline:
\begin{enumerate}
    \itemsep-.3em 
    \item \textbf{Pre-training.} Extract confidence predictor component after training a stand-alone detector on training data. 
    \item \textbf{Re-sampling.} Re-sample foreground (seen objects) and background bounding boxes in the training set so that they are equally populated;
    \item \textbf{Visual Feature Generation.} Train generator using the visual features of bounding boxes in (2.) and semantic data to synthesize visual features for unseen classes;
    \item \textbf{Confidence Predictor Re-training} Retrain confidence predictor with the real and synthetic visual features, and plug it back into the original detector.
\end{enumerate}

\noindent Following \cite{redmon2017yolo9000} we train YOLOv2 for Step 1. We describe the other steps in the sequel. 

\subsection{Foreground/Background Re-Sampling\label{subsec.gen_data}}
\noindent
Our objective is to construct a collection of visual features of (seen) foreground objects and background objects from the training set to reflect a balanced ratio of foreground/background objects. 
Note that the cell convolutional feature $F(I_m)$ is an inexpensive but effective visual representation of the bounding box proposals predicted by the current cell. However, not all cells are suitable to represent a bounding box since a cell may only overlap with the objects partially thus not sufficiently representative of the desired bounding box. Therefore, the re-sampled visual feature set $\mathcal{D}_{res}^{\alpha,\beta}$ (where $\alpha$ refers to the cell location and $\beta$ the bounding box index) is constructed as follows:

\noindent  {\bf Foreground}. For every image $I_m$, a cell feature $f$ is viewed as foreground if its associated bounding box prediction $\hat{\bf B}$ has a maximum Intersection over Union (IoU) greater than 0.5 on ground truth objects collections $\{\obj_m^{(i)}\}_{i=1}^{N_m}$, as well as its confidence prediction $\hat{p}_\conf$ is higher than 0.6. The feature $f$, along with its confidence score $\hat{p}_\conf$ and the ground truth class label $c$ of the object with maximum IoU, will be treated as a data point $(f, \hat{p}_\conf, c)$ \footnote{Other than the visual features, we also record the confidence value and the class label for reasons revealed in Sec~\ref{subsec.arc}}. 

\noindent {\bf Background}. A cell feature is viewed as background if its maximum IoUs over all ground truth objects is smaller than 0.2 and the associated confidence score is lower than 0.2. Features with top $r \times K_m$ smallest maximum IoU will be selected as the background data points $(f, \hat{p}_\conf, c_{bg})$, where $K_m$ is the number of foreground features extracted on image $I_m$, $r$ is the ratio of foreground/background data, and $c_{bg}$ is the class label for background which we set to -1. In our experiments, we let $r=1$ to balance the background and foreground objects.

In the sequel to avoid clutter, we omit the superscripts in $\mathcal{D}_{res}^{\alpha,\beta}$ and write the re-sampled visual feature set as  $\mathcal{D}_{res}$.

\subsection{Visual Feature Generation \label{subsec.arc}}
\begin{figure}
    \centering
    \includegraphics[width=.95\linewidth]{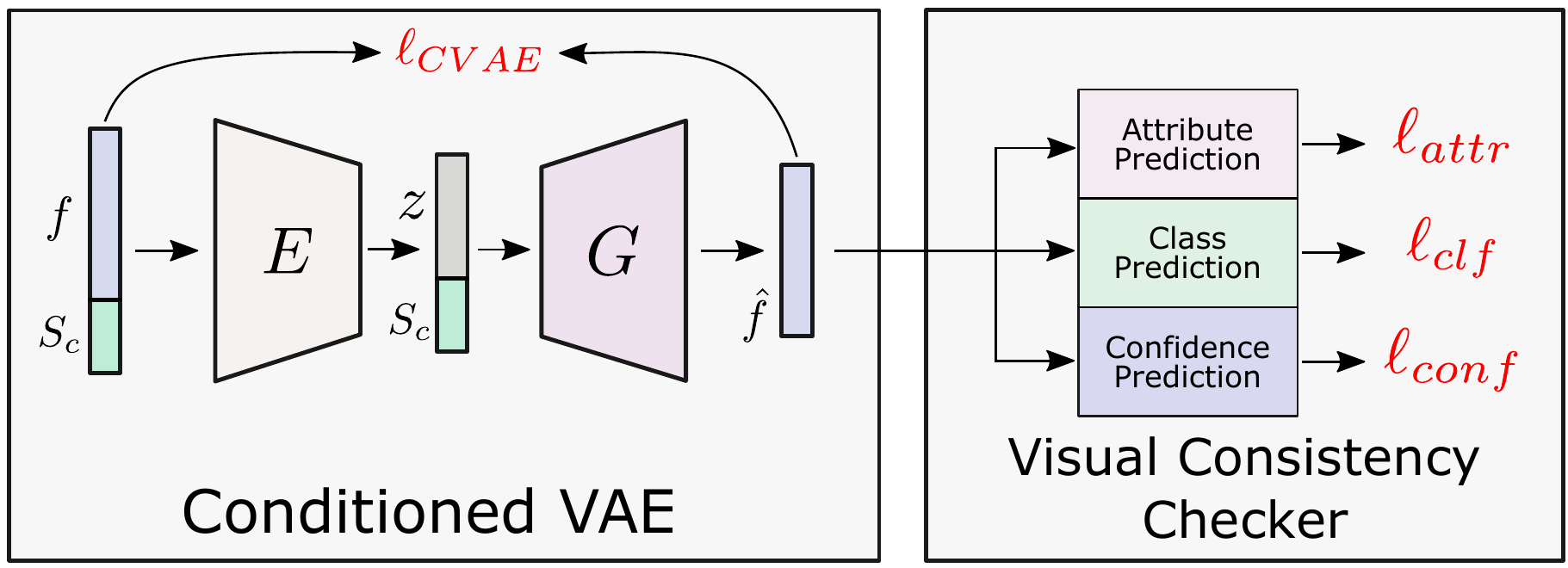}
    \caption{The proposed visual feature generator model.}
    \label{fig:cvae}
    \vspace{-5mm}
\end{figure}

\noindent 
After $\mathcal{D}_{res}$ is constructed, the next step is to train a visual feature generator to synthesize those features from their semantic counterparts while minimizing information loss. In particular, we construct our generator based on the concept of a conditional variational auto-encoder (CVAE) \cite{sohn2015learning}, but add an additional visual consistency checker component $D$ to provide more supervision, as shown in Fig.\ref{fig:cvae}. The CVAE is conditioned on the class semantic representation $S_c$, consisting of an encoder $E$ and a decoder $G$. The encoder takes the input as a concatenation of seen feature $f$ and semantic attribute $S_c$, and outputs the distribution of the latent variable $z$: $p_E(z|f,S_c)$. The decoder then generates exemplar feature $\hat{f}$ given the latent random variable $z$ and class semantic $S_c$: $\hat{f} = G(z, S_c)$. The visual consistency checker $D$ provides three additional supervisions on the generated feature $\hat{f}$ in addition to the reconstruction loss for CVAE: confidence consistency, attribute consistency loss, and classification consistency as described below.

\noindent \textbf{Conditional VAE}.
The decoder $G$ with parameters $\theta_G$, is responsible for generating unseen exemplar features which will be further used to retrain the confidence predictor. $\theta_G$ is trained along with $\theta_E$, the parameters of the encoder, by the conditional VAE loss function as following:
\begin{multline}
    \ell_{\mathrm{CVAE}}(\theta_G, \theta_E) = \mathrm{KL}\left( p_E(z|f,S_c)\|p(z) \right) \\
    - \mathbb{E}_{\mathcal{D}_{res}}\left[ \log p_G(f|z,S_c) \right]
\end{multline}
where the first term on right hand side is the KL divergence between the encoder posterior and prior of latent variable $z$, and the second term is the reconstruction error. Minimizing the KL divergence will enforce the conditional posterior distribution approximates the true prior. Following~\cite{kingma2013auto}, we utilize an isotropic multivariate Gaussian and the reparameterization trick to simplify these computations. 

\noindent \textbf{Visual Consistency Checker}.
This component provides multiple supervisions to encourage the generated visual features $\hat{f}$ to be consistent with the original feature $f$:

\noindent{\it \underline{Confidence Consistency}}: The reconstructed feature $\hat{f}$ should have the same confidence score as the original one, therefore, the confidence consistency loss is defined as the mean square error (MSE) between the confidence score of reconstructed and original features:
%
\begin{equation}
\itemsep-0.5em
    \ell_{\conf}(\theta_G) = \mathbb{E}_{\mathcal{D}_{res}}\left|\hat{p}_{\conf} - \Conf(\hat{f}) \right|^2 
\end{equation}
where  $\Conf(\cdot)$ refers to the confidence predictor model whose weights is frozen here for training the visual feature generator.

\noindent{\it \underline{Classification Consistency}}: The reconstructed feature $\hat{f}$ should also be discriminating enough to be recognized as the original category. Therefore, we feed $\hat{f}$ in to the classifier $\Clf$ and penalize the generator with the cross-entropy loss:
\begin{equation}
    \ell_{\clf}(\theta_G) = \mathbb{E}_{\mathcal{D}_{res}} [ \mathrm{CE}(\Clf(\hat{f}), c) ]
\end{equation}
where $c \in {\bf C}_{seen} \cup \{-1\}$ is the ground truth class for $f$, and $\Clf$ is pretrained by cross-entropy loss on $\mathcal{D}_{res}$ and will not be updated when training the generator. A class-weighted cross-entropy loss can also be used here to balance the data.

\noindent{\it \underline{Attribute Consistency}}: The generated feature should also be coherent with its class semantic.  We thus add an attribute consistency loss $\ell_{\attr}$ which back-propagates error to the generator between the attribute predicted on $\hat{f}$ and the conditioned class semantic:
\begin{equation}
    \ell_{\attr}(\theta_G) = \mathbb{E}_{\mathcal{D}_{res}} \left|S_c - \Attr(\hat{f}) \right|^2 
\end{equation}
where $S_{-1} = \mathbf{0}$ zero vector for background. The predictor $\Attr$ is also pretrained on $\mathcal{D}_{res}$ and the weights are frozen when optimizing $\ell_{\attr}$. Different class weights can also be applied for the purpose of data balance because the number of background is much larger than the other classes.

The parameters of CVAE can be end-to-end learned by minimizing the weighted sum of the CVAE and visual consistency checker loss functions:
\begin{equation}
    \theta_G^*, \theta_E^* = \argmin_{\theta_G, \theta_E} \ell_{\mathrm{CVAE}} + \lambda_{\conf} \cdot \ell_{\conf} + \lambda_{\clf} \cdot \ell_{\clf} + \lambda_{\attr} \cdot \ell_{\attr}
\end{equation}
where $\lambda_{[\cdot]}$ are the weights for the respective loss terms.

After the CVAE is trained, we can synthesize data feature for both seen and unseen objects by feeding the corresponding class attribute $S_c$ and latent variable $z$ randomly sampled from the prior distribution $p(z)$ to the decoder $G$. We generate $N_{seen}$ examples for every seen class and $N_{unseen}$ for every unseen class. We assume every synthesized data is ground truth object and assign 1 as its target confidence score, thus the synthesized data is constructed as $\mathcal{D}_{syn} = \{\hat{f}, 1, c\}$ where $c \in {\bf C}_{all}$.

\begin{figure*}[t]
    \includegraphics[width=0.5\linewidth]{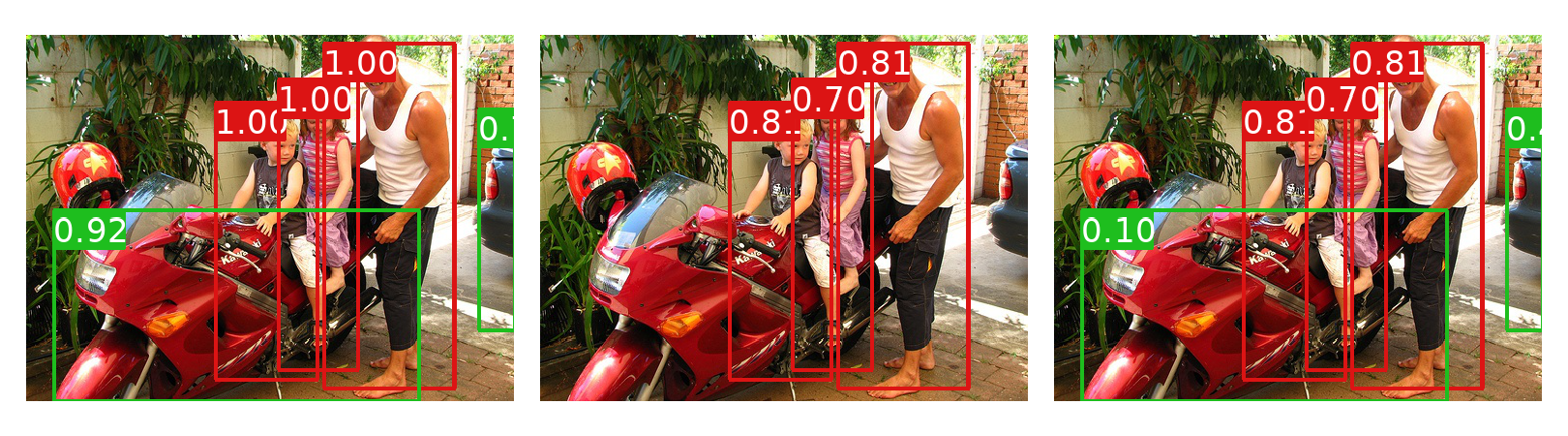}
    \includegraphics[width=0.5\linewidth]{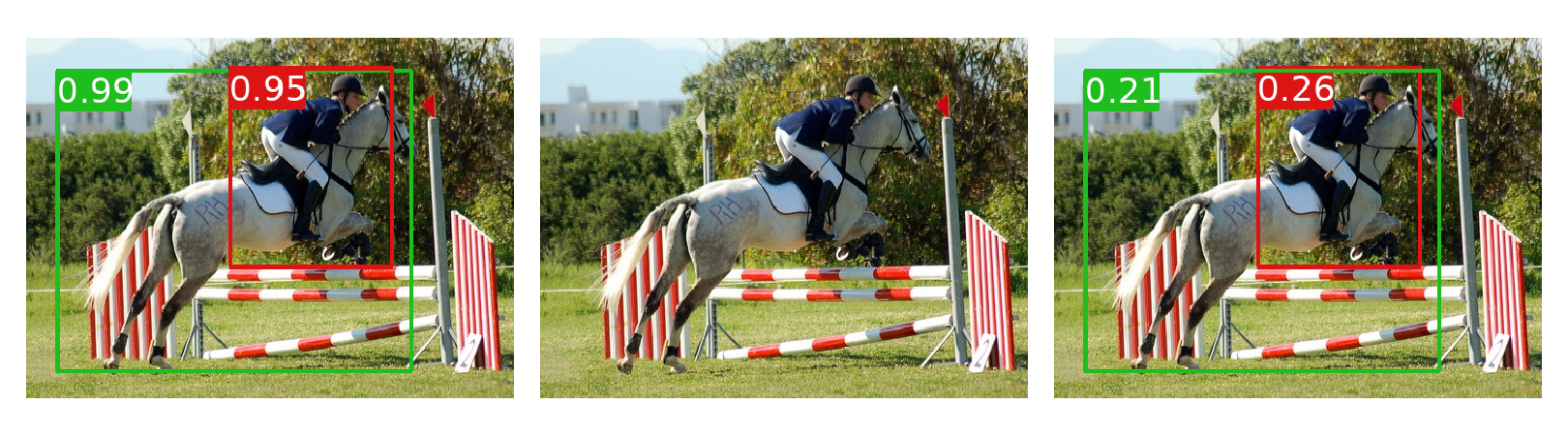}
    \includegraphics[width=0.5\linewidth]{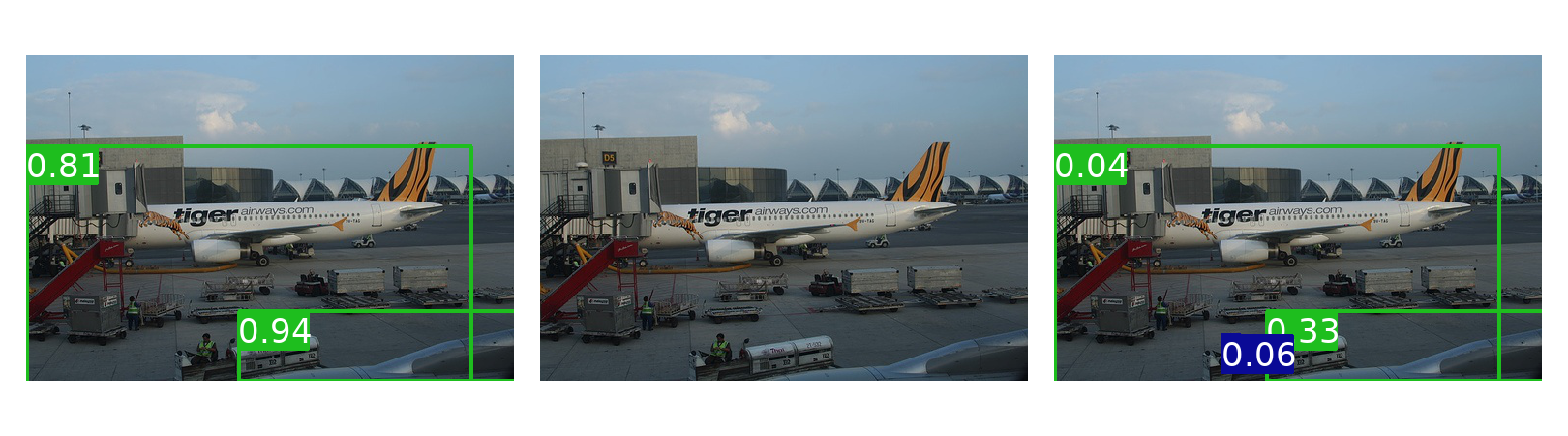}
    \includegraphics[width=0.5\linewidth]{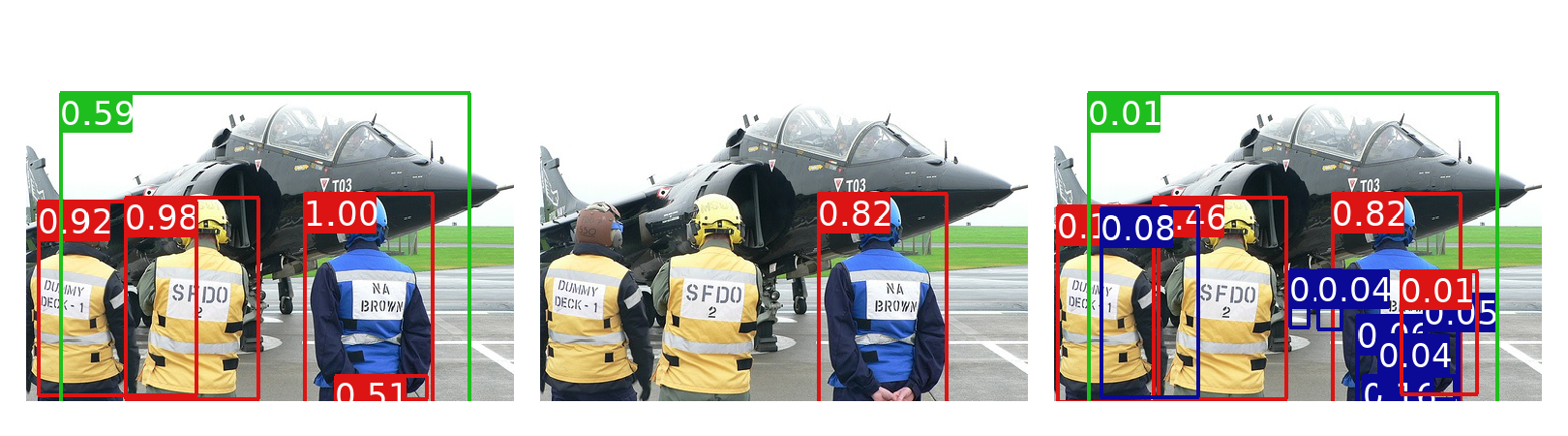}
    \includegraphics[width=0.5\linewidth]{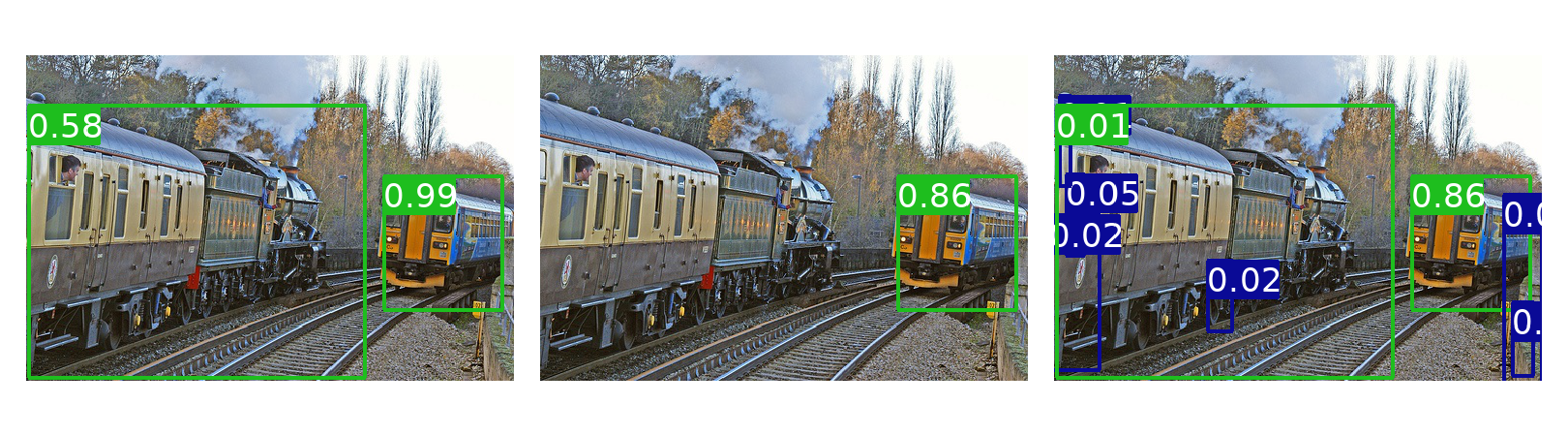}
    \includegraphics[width=0.5\linewidth]{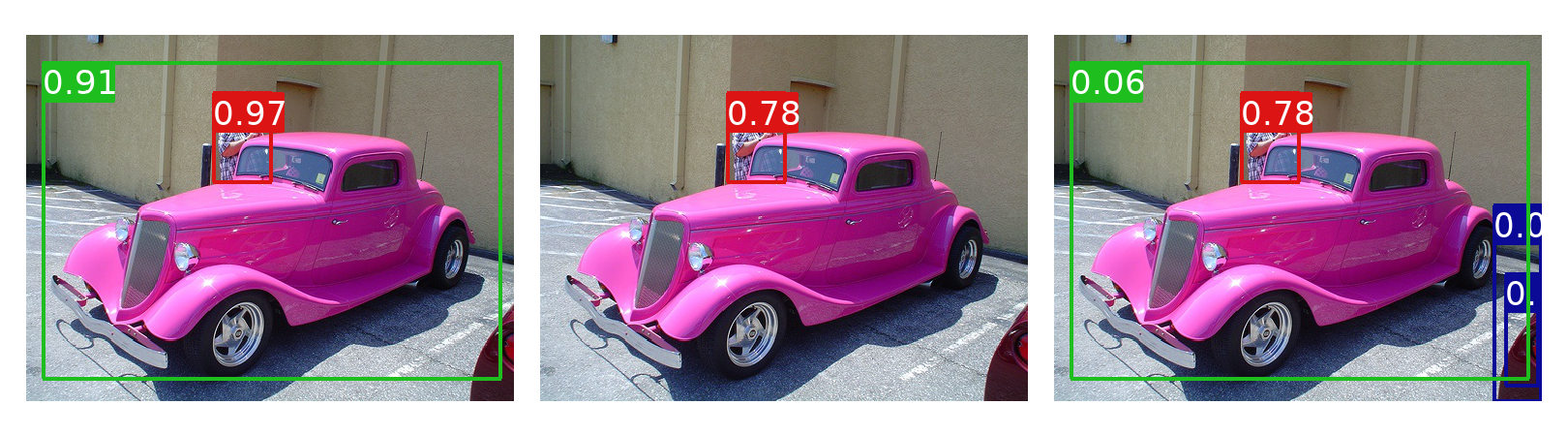}
    \includegraphics[width=0.5\linewidth]{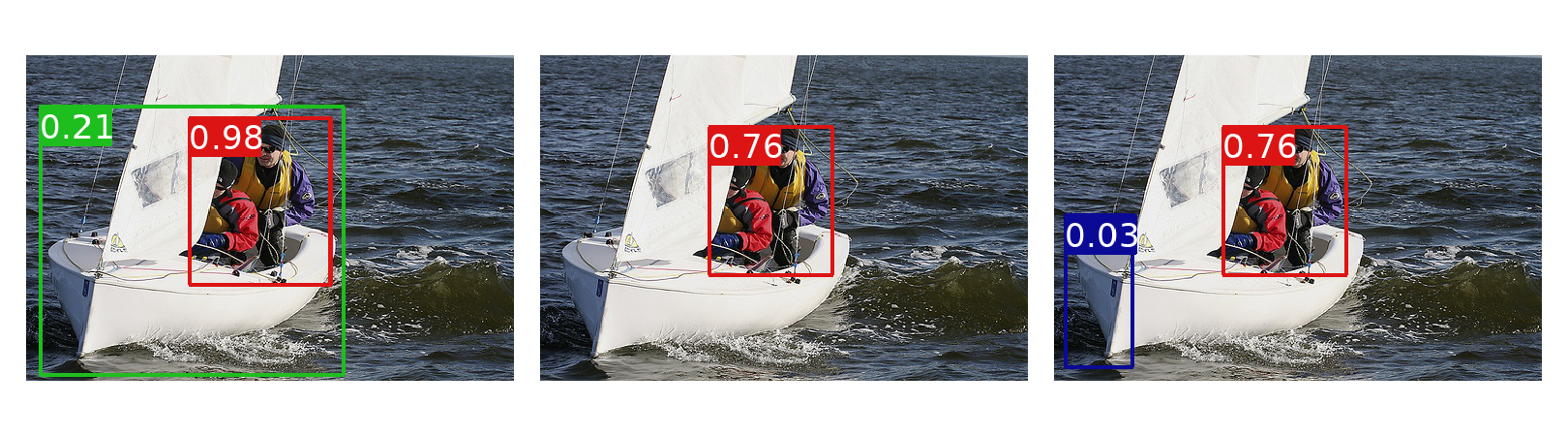}
    \includegraphics[width=0.5\linewidth]{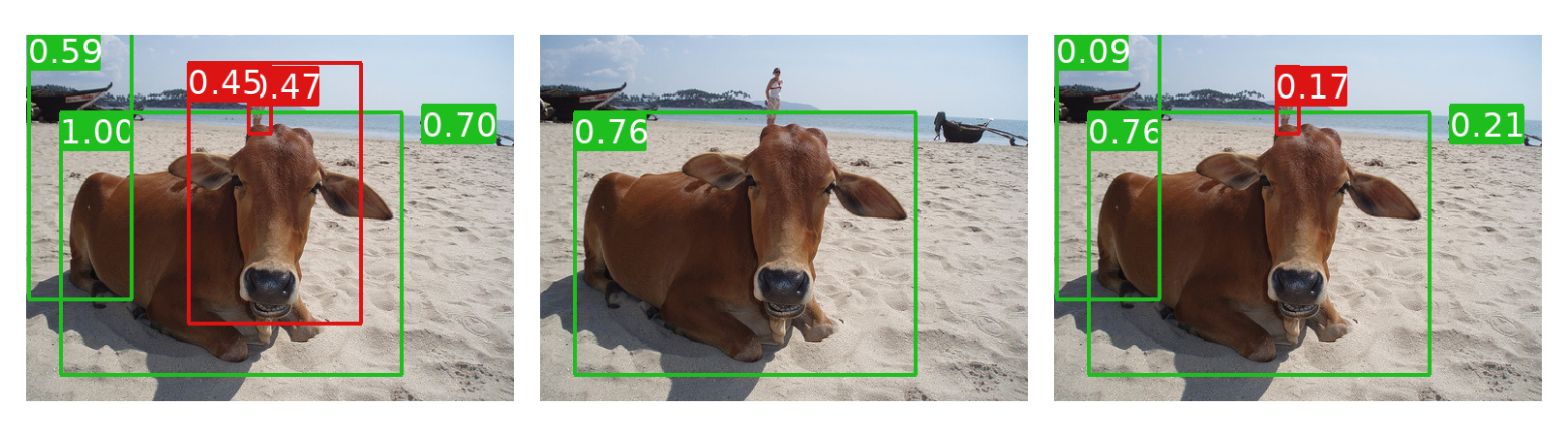}
    \caption{Visual examples of our ZSD results. Each triple shows: (from left to right) DELO detection results, vanilla YOLOv2 detection results at the same confidence threshold as DELO, vanilla YOLOv2 detection results at a much smaller confidence threshold. The seen, unseen and errors are color-coded as {\color{red} red}, {\color{green} green} and {\color{blue} blue}. Notice that compared to DELO, the vanilla YOLOv2 constantly predicts extremely low objectness scores on unseen objects, and suffers from significant detection errors for those unseen objects to be detected. }
    \label{fig:example}
    \vspace{-2mm}
\end{figure*}

\subsection{Confidence Predictor Re-Training}
With the collected real features $\mathcal{D}_{res}$ and the synthetic visual features $\mathcal{D}_{syn}$ which contains the generated visual features of unseen classes, 
we are now ready to re-train the confidence predictor $\Conf(\cdot)$ to encourage its confidence prediction on unseen objects while retaining its performance on seen and background objects.
Specifically, $\Conf(\cdot)$ is then re-trained on the combination of the extracted and synthetic features $\mathcal{D}_{res} \cup \mathcal{D}_{syn}$. Following the original YOLOv2\cite{redmon2017yolo9000}, MSE loss is used and the formal loss function is defined as:
\begin{multline}
    \ell = \frac{1}{|\mathcal{D}_{res}|} \sum_{f \in \mathcal{D}_{res}} |\Conf(f) - \hat{p}_{\conf}|^2 
    \\ + \frac{1}{|\mathcal{D}_{syn}|} \sum_{\hat{f} \in \mathcal{D}_{syn}} |\Conf(\hat{f}) - 1|^2
\end{multline}
We load weights from pretrained YOLOv2 in the first step (see Sec~\ref{subsec.procedure}) as a warm-start. Training with loss, $\ell$, encourages the confidence predictor to output higher scores for unseen object features while preserving existing confidences for seen objects and background.

\subsection{Implementation Details\label{subsec.imp_detail}}
The encoder $E$ and decoder $G$ are both two fc-layer networks in our CVAE model. The input size of $E$ is $N_{feat} + N_{attr}$ where $N_{feat} = 1024$ is the feature size and $N_{attr}$ is the length of the class semantic. The output size of $E$ which is also the size of latent variable $z$, $N_{latent}$, is set to 50. The input size of $G$ is $N_{latent} + N_{attr}$ and the hidden layer of $E$ and $G$ has 128 nodes.

For the visual consistency checker, both the classifier $\Clf(\cdot)$ and attribute predictor $\Attr(\cdot)$ are paramterized by a two FC-layer networks with hidden size 256. When pretrained on $\mathcal{D}_{res}$, $\Clf(\cdot)$ is trained 5 epochs with learning rate 1e-4 and $\Attr(\cdot)$ is trained 10 epochs with learning rate 1e-4, respectively.  
We set $\lambda_{\conf} = 1$, $\lambda_{\clf} = 2$ and $\lambda_{\attr} = 1$. We generate $N_{seen} = 50$ examples for every seen classes and $N_{unseen} = 1000$ for unseen.

\begin{table*}[t]
\vspace{-10mm}
\begin{minipage}[t]{0.6\textwidth}
\centering
    \renewcommand{\arraystretch}{1.15}
    \setlength{\tabcolsep}{2mm}
    \begin{tabular}{|l||l c c c||l c c c|}
        \hline
        \multirow{2}{*}{\bf Method} & \multicolumn{4}{c||}{\bf Pascal VOC} & \multicolumn{4}{c|}{\bf MS COCO} \\
        & \bf split & TU & TS & TM & \bf split & TU & TS & TM \\
        \hline
        YOLOv2 & \multirow{3}{*}{5/15} & 36.6 & 85.6 & 30.0 & \multirow{3}{*}{20/20} & 37.3 & 34.5 & 12.3 \\
        ZS-YOLO  & & 37.3 & 85.0 & 30.9 & & 40.6 & 41.2 & 20.2\\
        DELO & & \bf 39.4 & \bf 88.2 & \bf 34.7 & & \bf 41.5 & \bf 54.3 & \bf 41.6 \\
        \hline
        \hline
        YOLOv2 & \multirow{3}{*}{10/10} & 56.4 & 71.6 & 54.3 & \multirow{3}{*}{40/20} & 40.8 & 48.7 & 24.6\\
        ZS-YOLO & & 60.1 & 71.0 & 53.9 & & 42.7 & 44.0 & 30.0 \\
        DELO & & \bf 61.3 & \bf 73.5 & \bf 59.6 & & \bf 44.4 & \bf 49.7 & \bf 37.5\\
        \hline
        \hline
        YOLOv2 & \multirow{3}{*}{15/5} & 55.3 & 75.3 & 53.6 & \multirow{3}{*}{60/20} & 34.9 & 44.8 & 37.6\\
        ZS-YOLO & & 57.3 & 73.9 & 53.8 & & 43.8 & 40.6 & 33.6\\
        DELO & & \bf 58.1 & \bf 76.3 & \bf 58.2 & & \bf 48.9 & \bf 47.7 & \bf 39.4 \\
        \hline
    \end{tabular}
    \caption{Zero-shot detection evaluation results on various datasets and seen/unseen splits. TU = Test-Unseen, TS = Test-Seen, TM = Test-Mix represents different data configurations. Overall average precision (AP) in \% is reported. The highest AP for every setting is in {\bf bold}. }
    \label{tab:zsd_result}
\end{minipage}
\hspace{5mm}
\begin{minipage}[]{0.37\textwidth}
\centering
\renewcommand{\arraystretch}{1.5}
    \setlength{\tabcolsep}{1.2mm}
    \vspace{10mm}
    \begin{tabular}{|l|c c c|}
        \hline
        \bf Method & TU & TS & TM \\
        \hline
        YOLOv2 & 56.4 & 71.6 & 54.3 \\
        BS-1 & 59.5 ({\it 3.1}) & 73.2 ({\it 1.6}) & 58.5 (\it 4.2)\\
        BS-2 & 60.6 ({\it 4.2}) & 73.4 ({\it 1.8}) & 59.0 (\it 4.7)\\
        BS-3 & 61.0 ({\it 4.6})& 73.4 ({\it 1.8}) & 59.4 ({\it 5.1}) \\
        \hline
        DELO & 61.3 ({\bf\textit 4.9}) & 73.5 ({\bf\textit1.9}) & 59.6 ({\bf\textit 5.3})\\
        \hline
    \end{tabular}
    \vspace{3mm}
    \caption{Evaluation on the 10/10 split of Pascal VOC for baseline models. TU = Test-Unseen, TS = Test-Seen, TM = Test-Mix. Overall average precision in \% is reported. The difference between original YOLOv2 is reported in ($\cdot$) and the highest difference is in {\bf bold}.}
    \label{tab:contrib}
\end{minipage}
\vspace{-6mm}
\end{table*}

\section{Experiments\label{sec.exp}}





To evaluate the performance of our method, DELO, we conduct extensive qualitative and quantitative experiments. We tabulate results against other recent state-of-the-art methods, and then perform ablative analysis to identify important components of our system. 
We follow the protocol of \cite{zhu2019zero}, which emphasizes the need for evaluation both seen and unseen examples at test-time. As in \cite{zhu2019zero} we consider only visual seen examples during training. In summary, 
\textbf{(1)} Consider \textit{generalized ZSD setting} and omit results for the transductive generalized setting of \cite{rahman2019transductive} and somewhat de-emphasize the purely unseen detection results of \cite{bansal2018zero,rahman2018zero,li2019zero});
\textbf{(2)} Consider \textit{multiple splits with various seen/unseen ratios} in contrast to tabulating results for single splits with large seen/unseen ratios by \cite{rahman2018zero,bansal2018zero});
\textbf{(3)} consider \textit{multi-object image datasets}, and results for other datasets such as  F-MNIST that has clear black backgrounds as in~\cite{Demirel2018ZeroShotOD} or single objects/image such as ImageNet as in ~\cite{rahman2018zero,rahman2019transductive}). More detailed discussions can be found in Sec.~\ref{sec.intro} and Sec.~\ref{sec.related}. \\
%
%
%
\noindent{\bf Datasets.} We choose Pascal VOC \cite{pascal-voc-2012} and MSCOCO \cite{lin2014microsoft}, both of which are well known detection benchmarks, and as such exhibit multiple objects per image. PascalVOC has only 20 classes. For this reason, our goal here is to primarily understand how performance varies with different split ratios of seen to unseen objects (5/15, 10/10, and 15/5). MSCOCO is a larger dataset with about 80 classes and serves the purpose of understanding performance for fixed collection of unseen classes as the number of seen classes increase (20, 40 to 60). \\
\noindent{\bf Setting.} For each seen/unseen split, we evaluate our method on three data configurations: Test-Seen (TS), Test-Unseen (TU), and Test-Mix (TM) \cite{zhu2019zero}. For Test-seen our test images only contain objects from seen classes; test-unseen are images that only contain unseen objects; and test-mix are those that contain both seen and unseen objects. Test-mix is the generalized ZSD setting and is the most challenging, where the model needs to detect seen and unseen objects simultaneously. Following \cite{zhu2019zero}, we also use 0.5-IoU and 11 points average precision (AP) for evaluation.\\
\noindent{\bf Semantic Information.} Following \cite{zhu2019zero}, we use the attribute annotation from aPY \cite{farhadi2009describing} as the semantics on Pascal VOC. The semantic vectors are obtained by averaging the object-level attribute of all examples in the class. We use PCA to reduce dimensions to 20 to mitigate noise. On MSCOCO, a 25-dim word embedding w2vR proposed in \cite{zhu2019zero} is used. The semantics are re-scaled to $[0,1]$ on each dimension for both Pascal and MSCOCO.\\
\noindent {\bf Training Details}
For Pascal VOC, CVAE is trained by an Adam optimizer with a learning rate of $1e-4$. On the 10/10 and 15/5 splits, we set training epochs to 60 and scale the learning rate by 0.5 every 15 epochs. On the 5/15 split, the training epoch is 200 and the learning rate is scaled by 0.5 every 60 epochs. 
On MSCOCO, the learning rate is set to $1e-4$. On 20/20 split the model is trained for 60 epochs, while on 40/20 and 60/20 splits, the model is trained for 40 epochs. The learning rate is scaled by 0.5 every 15 epochs.







\subsection{Zero Shot Detection Evaluation\label{subsec.zsd_eval}}
\noindent 
\textbf{Tabulating AP.} We evaluate DELO on all seen/unseen splits as well as Test-Seen/Unseen/Mix configurations (Table~~\ref{tab:zsd_result}) against vanilla YOLOv2 \cite{redmon2017yolo9000} trained in a standard fully-supervised manner with the training partition, as well as the state-of-the-art ZSD method ZS-YOLO \cite{zhu2019zero}. 

\noindent \textbf{Discussion Part-A}. \\
(1) {\it Vanilla YOLOv2 does well on Test-Seen}.
The state-of-art YOLOv2 as reported in \cite{redmon2017yolo9000} are 73.4\% mAP on Pascal VOC2012 and 44.0\% mAP on MSCOCO. Observe from Table~~\ref{tab:zsd_result} that the vanilla YOLOv2 trained on seen partition achieves similar performances on Test-Seen with no unseen objects, i.e. 85.6\%, 71.6\%, 75.3\%  for Pascal VOC (5/15. 10/10. 15/5 split, respectively), and 34.5\%, 48.7\%, 44.8\% for MS COCO (20/20, 40/20, 60/20 split, respectively). Consequently, YOLOv2 is a strong baseline to compare against particularly for test-seen. Furthermore, as we increase the split ratio the number of seen classes increases, and consequently, test-mix tends to favor seen class detection. For this reason we should expect vanilla YOLOv2 detector to perform better in this case as well.\\
\noindent (2) {\it Re-training with synthetic visual features improves detection performance}. DELO consistently outperforms vanilla YOLOv2 and ZS-YOLO on all test configurations. ZS-YOLO uses semantic features to train the confidence predictor, which can be visually noisy (attributes such as ``useful''). As a result, while improving upon YOLOv2 on Test-Unseen/-Mix, its Test-seen, performance is lower, e.g. MS COCO 40/20 split, ZS-YOLO gets 44.0\% on Test-Seen, compared to YOLOv2's 48.7\%. In contrast, DELO's confidence predictor leverages visual features from seen/unseen/background boxes. Additionally, the feature pool is re-sampled according to a more balanced foreground/background ratio. Consequently, DELO, also improves Test-Seen performance, e.g. MS COCO all splits, we see average DELO's AP is (2.53\% / 8.63\% / 11.53\%) and (7.25\% / 7.90\% / 14.63\%)  better than ZS-YOLO and YOLOv2 respectively, on Test-Unseen/Seen/Mix.\\
\noindent (3) {\it DELO is robust to different seen/unseen configurations}. YOLOv2 and ZS-YOLO's performance changes significantly with large number of classes (MSCOCO). As seen classes increases and unseen classes remain the same, YOLOv2 realizes (12.3\% / 24.6\% / 37.6\%) on Test-Mix; ZS-YOLO realizes (20.2\% / 30.0\% / 33.6\%). Compared to these, DELO produces a much more consistent detection performance (41.6\% / 37.5\% / 39.4\%). On Pascal VOC, performance of all the three models varies significantly for different splits of Test-Mix since the dataset is of a smaller scale and the number of unseen classes is also changing. But DELO's performance is still superior. \\ 
%
\begin{table}[t]
    \centering
    \renewcommand{\arraystretch}{1.2}
    \setlength{\tabcolsep}{1.1mm}
    \begin{tabular}{|l|c c|c c|}
        \hline
        \multirow{2}{*}{\bf Method} & \multicolumn{2}{c|}{Recall@100} & \multicolumn{2}{c|}{mAP}\\
        & ZSD & GZSD & ZSD & GZSD \\
        \hline
        SB\cite{bansal2018zero} &  24.4 & - & 0.70 & - \\
        DSES\cite{bansal2018zero} &  27.2 & 15.2 & 0.54 & -\\
        TD\cite{li2019zero} &  34.3 & - & - & - \\
        YOLOv2 &  24.8 (51.6) & 30.8 (52.8) & 5.4 & 9.6\\
        DELO &  33.5 (55.7) & 36.6 (55.8) & 7.6 & 13.0 \\
        \hline
    \end{tabular}
    \caption{ZSD and GZSD performance evaluated with Recall@100 and mAP on MS COCO to compare with other ZSD methods. A 2-FC classifier trained on $\mathcal{D}_{syn}$ is appended to YOLOv2 and DELO to conduct the full detection. The number in the parenthesis is class-agnostic recall ignoring classification.}
    \label{tab:R100}
    \vspace{-6mm}
\end{table}

\noindent \textbf{Tabulating Recall@100 and mAP.} 
Following the protocol in~\cite{bansal2018zero}, we conducted a second set of experiments on MS COCO, adopting the Recall@100 and mAP as evaluation metrics, to baseline against~\cite{bansal2018zero, li2019zero} (more details in Sec.~\ref{sec.related}). The configurations and ZSD performance are reported in Table~\ref{tab:R100}.

\noindent \textbf{Discussion Part-B}.\\
\noindent \textit{ZSD is in essence a classification problem under the Recall@100 metric at high seen/unseen ratio}. Observe that, a vanilla detector, e.g. YOLOv2 in cascade with a ZSR model (we chose a 2-FC classifier trained on $\mathcal{D}_{syn}$) achieves similar performance on Recall@100 as existing ZSD methods, i.e.  24.8 (YOLOv2) compared to 24.4 (SB~\cite{bansal2018zero}), 27.2 (DSES~\cite{bansal2018zero}), 34.3 (~\cite{li2019zero}). 
Fundamentally, the issue is two-fold. First at large split ratio's the current methods benefit from unseen visual features that resemble seen examples, and so do not require better detections. Vanilla detectors that are not optimized for unseen objects are capable of localizing unseen objects. Second, the Recall@100 metric also helps in this process since 100 bounding boxes typically contain all unseen objects at the high split ratios. Once this is guaranteed, background boxes can be eliminated based on post-processing with a zero-shot classifier that rejects background whenever no unseen class is deemed favorable. For this reason, we also present other metrics such as AP in Table~\ref{tab:zsd_result} as well as (in Table~\ref{tab:R100} within brackets) whether bounding boxes are true objects. In addition we see that mAP improves both under ZSD as well as the more important GZSD setting. Finally, while TD is marginally better on ZSD, we emphasize that it is possible to bias ZSR models towards unseen classes when we are cognizant of the fact that no seen classes are present~\cite{xian2018zero}. Note that a large fraction of our bounding boxes are indeed correct, and so our lower performance can be attributed to the fact that we did not fine tune our ZSR model.

\subsection{Ablative Analysis\label{subsec.AA}}
\noindent \textbf{Contribution of visual consistency checker.}
The visual consistency checker $D$ in our generative model provides more supervision to the decoder to encourage it generates better exemplar features. To measure the contribution of each components in the visual consistency checker, we compare with three baselines:
{\bf (1) BS-1}: the entire visual consistency checker $D$ is removed, the model is thus reduced to a standard CVAE and trained only by $\ell_{\mathrm{CVAE}}$.
{\bf (2) BS-2}: only the confidence predictor is used in the visual consistency checker and the model is trained by $\ell_{\mathrm{CVAE}} + \lambda_{\conf}\cdot\ell_{\conf}$
{\bf (3) BS-3}: the attribute predictor is removed from the visual consistency checker and the model is trained by $\ell_{\mathrm{CVAE}} + \lambda_{\conf}\cdot\ell_{\conf} + \lambda_{\clf}\cdot\ell_{\clf}$.
We evaluate the baseline models on 10/10 split of Pascal VOC and the results along with the differences between the original YOLOv2 are tabulated in Table.\ref{tab:contrib}.

It is apparent that with all the visual consistency checker components included, DELO realizes optimal performance. Without any supervision from the consistency checker, the pure CVAE achieves 59.5 on TU, 73.2 on TS and 58.5 on TM (BS-1). Incorporating the confidence predictor $\Conf$ increases 1.1 on TU and 0.5 on TM, and the classifier $\Clf$ contributes 0.3 improvement on TU and 0.4 on TM. Finally, by integrating the attribute predictor, the performance further increases 0.3 on TU, 0.1 on TS, and 0.2 on TM. The visual consistency checker improves the overall performance, especially on TU and TM, as it encourages the generated data to be more consistent to the original feature and the class information.

\begin{figure}
    \centering
    \includegraphics[width=.95\linewidth]{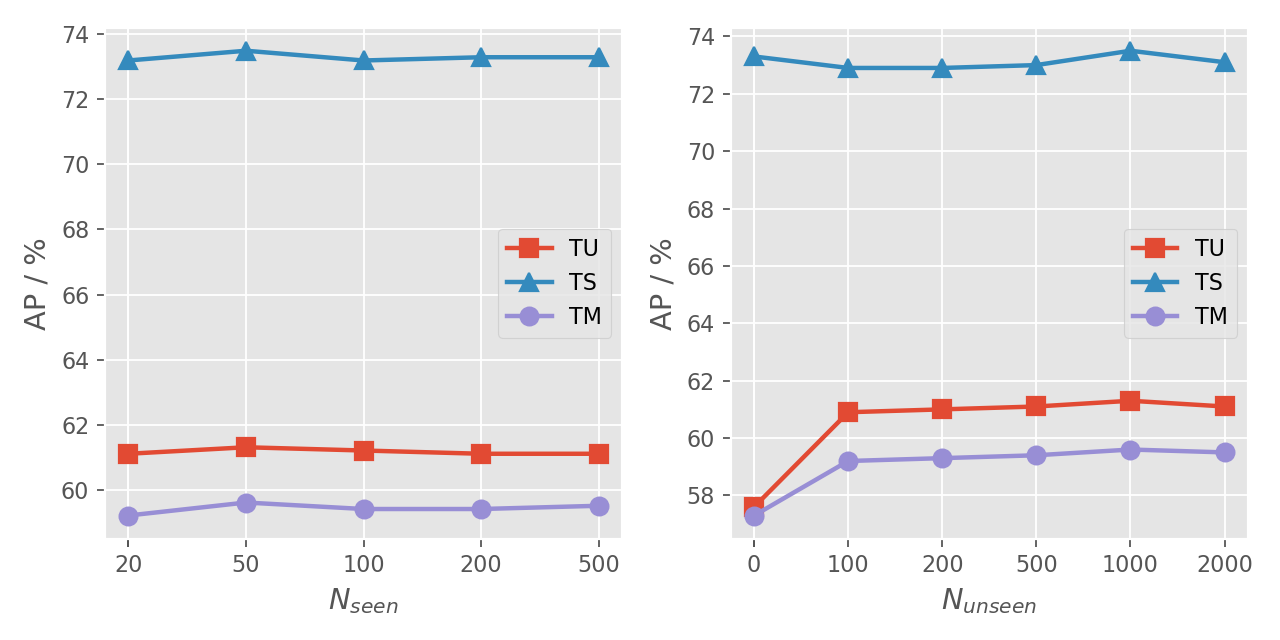}
    \caption{Performance of various $N_{seen}$ (left) and $N_{unseen}$ (right) on 10/10 split of Paascal VOC. TU = Test-Unseen, TS = Test-Seen, TM = Test-Mix.}
    \label{fig:vary_number}
    \vspace{-6mm}
\end{figure}

\noindent \textbf{Number of Generated Examples.}
We also perform experiments to evaluate how the number of generated examples affects the detection performance. In the experiment, we first vary $N_{seen}$ in the range $[20, 50, 100, 200, 500]$ while keeping $N_{unseen} = 1000$. Then we vary $N_{unseen}$ in the range of $[0, 100, 200, 500, 1000, 2000]$ while set $N_{seen} = 50$. 
The experiments are conducted on 10/10 split of Pascal VOC and the final detection performance are visualized in Fig.\ref{fig:vary_number}.
The generated unseen data plays an important role in the method, as we can see the performance on TU and TM drops $>2\%$ when training with $N_{unseen} = 0$. The performance on TU and TM increases when more unseen data are available, and get saturated after $N_{unseen} > 1000$. A small number of unseen examples (e.g. 100) is sufficient for learning a strong confidence predictor. The number of seen generated data, on the other hand, only affects the overall performance slightly as it has similar distribution as $\mathcal{D}_{res}$


\section{Conclusion\label{sec.conclusion}}
\noindent
We proposed DELO, a novel Zero-shot detection algorithm for localizing seen and unseen objects. We focus on the generalized ZSD problem where both seen and unseen objects can be present at test-time, but we are only provided examples of seen objects during training. Our key insight is that, while vanilla DNN detectors are capable of producing bounding boxes on unseen objects, these get filtered out due to poor confidence. To address this issue DELO synthesizes unseen class visual features, leveraging semantic data. Then a confidence predictor is trained with training data augmented with synthetic features. We employ a conditional variational encoder, with additional loss functions, that are specifically chosen to improve detection performance. We also propose a re-sampling strategy to improve the foreground/background during training. Our results show that on a number metrics, on complex datasets involving multiple objects/image, DELO achieves state-of-the-art performance.

\section*{Acknowledgement}
This work was supported partly by the National Science Foundation Grant  1527618, the Office of Naval Research Grant N0014-18-1-2257and by a gift from ARM corporation. 

{\small
\bibliographystyle{ieee_fullname}
\bibliography{main}
}





\end{document}